\documentclass{article}
\usepackage{spconf,amsmath,graphicx}

\usepackage{amsmath}
\usepackage{tipa}
\usepackage{url}

\title{Measuring Sound Symbolism in Audio-Visual Models}
\name{Wei-Cheng Tseng*, Yi-Jen Shih\sthanks{equal contribution}, David Harwath, Raymond Mooney}
\address{The University of Texas at Austin}
\usepackage{multirow}
\usepackage{multicol}
\usepackage{amssymb}
\usepackage{xcolor}

\begin{document}
%
\maketitle
\begin{abstract}
Audio-visual pre-trained models have gained substantial attention recently and demonstrated superior performance on various audio-visual tasks. This study investigates whether pre-trained audio-visual models demonstrate non-arbitrary associations between sounds and visual representations-known as sound symbolism--which is also observed in humans. We developed a specialized dataset with synthesized images and audio samples and assessed these models using a non-parametric approach in a zero-shot setting. Our findings reveal a significant correlation between the models' outputs and established patterns of sound symbolism, particularly in models trained on speech data. These results suggest that such models can capture sound-meaning connections akin to human language processing, providing insights into both cognitive architectures and machine learning strategies.
\end{abstract}

\section{Introduction}
The field of deep learning has sustained significant interest in exploring the interplay between auditory and visual perceptions, two crucial modalities of human sensory processing. Known as audio-visual learning~\cite{zhu2021deep}, this research field seeks to address the limitations of single-modality learning, enhancing performance on existing applications while opening new avenues for research. A key area of focus within audio-visual learning is representation learning, which aims to develop integrated audio-visual representations. Methods in this area often involve pretraining models to maximize the mutual information between auditory and visual inputs, whether in image or video form.

By training on diverse audio-visual datasets using specific algorithms, these models develop various specializations, such as speech-image retrieval~\cite{shihSpeechCLIP, peng2022fastvgs}, visual speech recognition~\cite{shi2022avhubert}, and audio event understanding~\cite{gong2023cavmae, huang_mavil}. These pre-trained models can extract highly informative, modality-agnostic representations from raw data, making them highly adaptable for audio-visual tasks, even with limited labeled data.

Beyond these specializations, some models have demonstrated intriguing language-processing capabilities. For instance,~\cite{shi2022avhubert} shows that learning from speech with accompanying lip recordings enhances acoustic unit discovery. Additionally,~\cite{Harwath_2018_ECCV, peng2022word} demonstrates that audio-visual models trained solely on images paired with spoken captions, in a manner similar to human language acquisition, improves word discovery and segmentation abilities. Building on this,~\cite{lai2023audio} further explored the potential of these models to perform syntactic parsing of spoken sentences.

\begin{figure}[!t]
    \centering    
    \vspace{3pt}
    \includegraphics[width=0.95\linewidth]{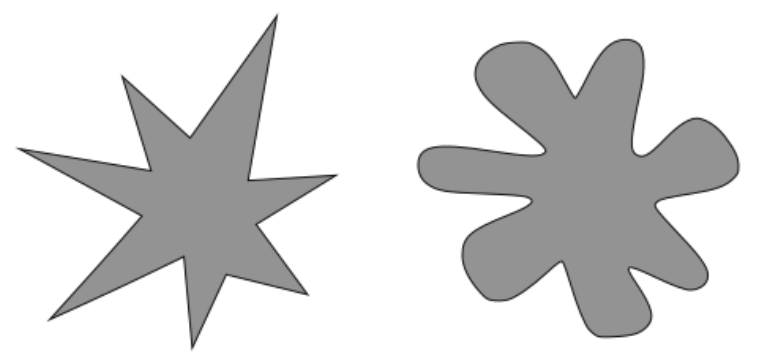}
    \vspace{3pt}
    \caption{Example of the kiki-bouba experiment: When hearing the names "kiki" and "bouba", people from various cultural and linguistic backgrounds typically label the left shape as "kiki" and the right one as "bouba".}
    \label{fig:enter-label}
    
\end{figure}
Building on these advancements, an intriguing research frontier lies in understanding how these models handle more abstract linguistic phenomena, such as sound symbolism~\cite{bredin1996onomatopoeia, firth1968tongues, kohler1967gestalt, Synaesthesia}. Sound symbolism refers to the systematic and non-arbitrary association between sounds and meanings—a phenomenon that challenges the traditional view of linguistic arbitrariness~\cite{locke1847essay}, suggesting that certain sounds naturally evoke specific meanings or visual qualities. For instance, the widely studied "kiki-bouba effect," illustrated in fig~\ref{fig:enter-label}, demonstrates that people tend to associate the sharp-sounding word "kiki" with angular shapes, while the softer-sounding "bouba" is linked to rounder shapes~\cite{kohler1967gestalt, Synaesthesia}. This phenomenon is robust across cultures and age groups~\cite{kohler1967gestalt, bremner2013bouba}, indicating that some sound-meaning associations may be universally grounded in human perception.

According to the sound symbolism bootstrapping hypothesis for language acquisition~\cite{imai2014sound}, sound symbolism may play a foundational role in language learning by aiding toddlers in mapping and integrating multi-modal inputs, thereby supporting the formation of lexical representations. Since audio-visual models share certain parallels with human learning processes, we hypothesize that these models may naturally capture such associations. By investigating sound symbolism in these models, we can not only deepen our understanding of the cognitive parallels between human perception and artificial learning systems but also potentially enhance the models' interpretability and alignment with human-like semantics in multimodal contexts. This exploration may thus open new directions for both linguistic theory and practical applications in audio-visual technology.

To explore this, we propose examining sound symbolism within audio-visual models by testing their ability to associate specific sounds with visual characteristics in ways consistent with human perception, as in the kiki-bouba effect. We start with curating a specialized dataset that includes synthesized images and audio samples, each distinctly categorized as "sharp" or "round." Following~\cite{alper2023kiki}, we then use a non-parametric approach in a zero-shot setting to probe the inherent knowledge of these models. Our experimental results reveal a significant correlation between the models' outputs and established patterns of the kiki-bouba effect in some audio-visual models. Specifically, audio-visual models trained on spoken image captions are able to group visual stimuli and audio with different appearances significantly better than chance. Additionally, audio-visual models generally show a more pronounced sound symbolism effect than their purely textual vision-language model counterparts. Our findings show synergy with existing psychological literature~\cite{article, monaghan2014arbitrary, Synaesthesia, ShapeOfBoubas, FletcherVisual} and further support the non-arbitrariness of human language.

\section{Related Work}
\subsection{Sound Symbolism}
Evidence of sound symbolism in human language has been extensively studied in the past few decades. In short, the findings suggest that linguistic sounds can not only be perceived as similar to natural sounds ~\cite{bredin1996onomatopoeia}, but also as similar to visual shape~\cite{kohler1967gestalt, Synaesthesia}, action~\cite{dingemanse2019ideophone}, magnitude~\cite{winter2021size}, abstract concepts~\cite{nichols1996amerind}, and language abstractions~\cite{firth1968tongues}. These associations, which demonstrate that sounds can convey meanings, challenge the arbitrariness assumption in semiotics. 
Furthermore, such evidence has been found in different languages and cultures~\cite{blasi2016sound, bremner2013bouba, cwiek2022bouba, joo2020phonosemantic}, implying a fundamental role of sound symbolism in language systems. Recent research also suggests the importance of sound symbolism for infants in building basic vocabulary and semantic clusters of lexicon, as well as in the evolution of language itself~\cite{imai2014sound}. Additionally, sound symbolism has been widely used in commerce for better naming of new brands~\cite{klink2000creating}.
\subsection{Pre-trained Audio-visual Models}
Audio-visual representation learning has emerged as a prominent research area in recent years~\cite{shihSpeechCLIP,shi2022avhubert,gong2023cavmae,huang_mavil,Girdhar2023Imagebind,peng2022fastvgs}. These techniques typically involve pre-training models using either masked autoencoding~\cite{he2022masked} or contrastive learning~\cite{oord2018representation}, and the knowledge acquired by these models is significantly influenced by the domain of the pre-training data.
For instance, AV-HuBERT~\cite{shi2022avhubert} focuses on visual speech recognition, using videos of lip movements to help the model associate visual cues with spoken language. 
Visually-grounded speech models, such as SpeechCLIP~\cite{shihSpeechCLIP} and FaST-VGS~\cite{ peng2022fastvgs}, are trained on paired images and spoken captions, enabling the development of associations between spoken words and visual content. 
Conversely, models trained on more general audio events~\cite{gong2023cavmae, huang_mavil}, such as videos of dogs barking, tend to learn object-sound correspondence and understand acoustic scenes.
These models are capable of extracting highly informative, modality-agnostic features from raw auditory and visual inputs.
The extracted features have been demonstrated to achieve great performance across various audio-visual tasks, including sound localization~\cite{murdock_avlocalization}, segmentation~\cite{Shentong_segmentation}, sound event classification, and audio-visual speech recognition~\cite{tseng2023avsuperb}.

Despite significant advances in pre-trained audio-visual models, the link between these models and sound symbolism has largely been overlooked in the literature. 
Given that humans reliably demonstrate sound symbolism effects, it is intriguing to explore whether this effect also emerges in models trained on multimodal data. 
Such research can bridge deep learning approaches in speech processing with cognitive science, offering insights into model behavior that aligns with human perceptual and cognitive process—an area of broad interest within the research community~\cite{caucheteux2023evidence,millet2022toward}.

\subsection{Interpreting Deep Learning Models}
Deep learning has undoubtedly achieved remarkable success across various fields~\cite{10.1145/3234150, DONG2021100379, alzubaidi2021review}. Researchers are increasingly interested in interpreting these models to better understand the underlying mechanics and rationales behind their decisions~\cite{BlackboxNLP}. This interpretation often involves feature visualization~\cite{erhan2009visualizing, olah2017feature, DBLP:journals/corr/ZeilerF13}, saliency methods~\cite{DBLP:journals/corr/SundararajanTY17, bach2015pixel, DBLP:journals/corr/ZeilerF13}, and probing tasks~\cite{tenney2019you, hewitt-manning-2019-structural, DBLP:journals/corr/MahendranV14, pasad2021layer}. Recently, the natural language processing community has shown a growing interest in comparing the behavior of vision-language models to aspects of human language processing through probing tasks. 
These tasks assess models on compositionality~\cite{thrush2022winoground}, abstraction~\cite{ji2022abstract}, and sound symbolism~\cite{alper2023kiki}.

In this work, we specifically aim to study the effect of sound symbolism in audio-visual models. By integrating insights from the interpretability of deep learning models, we explore how these black-box systems may encode and reflect non-arbitrary associations between sounds and visual representations, akin to human perceptual and cognitive processes.

\section{Dataset Collection}
To assess the presence of sound symbolism in audio-visual models, we need to inspect how models behave with respect to sound and visual stimuli categorized into sharp or round groups. Therefore, we first construct a specialized dataset with images and audios grounded and grouped by human preference.
While it is feasible to use controlled examples from original kiki-bouba experiments (as illustrated in Figure 1), the limited diversity in these samples could potentially undermine the reliability of our evaluation results. Hence, we enhanced the diversity by synthesizing data with generative models, as detailed in Sections 3.1 and 3.2. Overall, our dataset consists of 500 images and 3888 audio samples, each categorized into sharp or round groups, reflecting the classifications used in the original kiki-bouba experiment.

\subsection{Generating Images}
To generate images, we first defined two sets of adjectives reflecting object appearance, $\mathcal{W}_{\text{round}}$ for roundness and $\mathcal{W}_{\text{sharp}}$ for sharpness:
\begin{align*}
\mathcal{W}_{\text{round}} &= \{round, circular, soft, fat, chubby, \\ & curved, smooth, plush, plump, rotund\} \\
\mathcal{W}_{\text{sharp}} &= \{sharp, spiky, angular, jagged, hard, \\ & edgy, pointed, prickly, rugged, uneven\}
\end{align*}
Then, we prompted a state-of the art text-to-image model\footnote{https://huggingface.co/stabilityai/stable-diffusion-2-1} using the template $\mathcal{P}$, where $\langle w \rangle \in \mathcal{W}_{\text{round}}\cup\mathcal{W}_{\text{sharp}}$ is the adjective:
\begin{align*}
\mathcal{P}: \text{``A 3D-rendering of a } \langle w \rangle \text{ object''}
\end{align*}
For each adjective, we generated 25 images with different random seeds. This resulted in two collections of images, $\mathcal{I}_{\text{round}}$ and $\mathcal{I}_{\text{sharp}}$, each comprising 250 images. Figure 2 displays several examples from these generated images.

\begin{figure}
    \centering
    \includegraphics[width=0.9\linewidth]{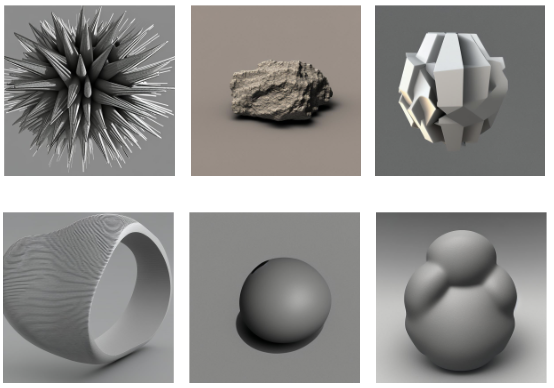}
    \caption{Examples of generated images: the upper row is from the sharp image set $\mathcal{I}_{\text{sharp}}$, while the lower row is from the round image set $\mathcal{I}_{\text{round}}$.}
    \label{fig:example_generatedImg}
    \vspace{-5pt}
\end{figure}

\subsection{Synthesizing Audios}
Similar to the image generation process, we aim to collect two sets of audio samples, $\mathcal{S}_{\text{round}}$ for round sounds and $\mathcal{S}_{\text{sharp}}$ for sharp sounds. We started with categorizing English consonants and vowels into round sounding or sharp sounding groups based on human preference alignments from prior work~\cite{mccormick2015sound}:
\begin{align*}
\mathrm{C}_{\text{round}} &= \{\text{\textipa{m}}, \text{\textipa{n}}, \text{\textipa{l}}, \text{\textipa{b}}, \text{\textipa{d}}, \text{\textipa{g}}\} &&\mathrm{V}_{\text{round}} = \{\text{\textipa{o:}}, \text{\textipa{u:}}\}\\
\mathrm{C}_{\text{sharp}} &= \{\text{\textipa{k}}, \text{\textipa{t}}, \text{\textipa{p}}, \text{\textteshlig}, \text{\textdyoghlig}, \text{\textipa{z}}\}\  &&\mathrm{V}_{\text{sharp}} = \{\text{\textipa{E}}, \text{\textipa{i:}}\}
\end{align*}
In addition, we also identified a set of neutral phones that shows insignificant tendency toward round or sharp.
\begin{align*}
\mathrm{C}_{\text{neutral}} &= \{\text{\textipa{f}}, \text{\textipa{s}}, \text{\textipa{v}}\}\  &&\mathrm{V}_{\text{neutral}} = \{\text{\textipa{a:}}\}
\end{align*}
We then create phone sequences using the three-syllable template $(\mathrm{C}\mathrm{V})_1(\mathrm{C}\mathrm{V})_2(\mathrm{C}\mathrm{V})_3$ with specific rules, as suggested by~\cite{alper2023kiki}: (1) the first and last syllable must be identical; (2) all phone sequences should not mix elements from the round group $\mathrm{C}_{\text{round}}\cup\mathrm{V}_{\text{round}}$ and the sharp group $\mathrm{C}_{\text{sharp}}\cup\mathrm{V}_{\text{sharp}}$. For example, a sequence like $\text{\textipa{[ki:mu:ki:]}}$ is not acceptable; (3) the initial phoneme must be drawn from either $\mathrm{C}_{\text{round}}$ or $\mathrm{C}_{\text{sharp}}$, to prevent sequences composed entirely of neutral phones. This design of the template helps ensure that the sequences are less likely to resemble existing words, thus avoiding potential memorization from the pre-training and focus only on the phonetic characteristics of these sounds.
Several examples of valid phone sequence are shown as followings:
\begin{align*}
    round&: \text{\textipa{[mu:lu:mu:]}}\ \  \text{\textipa{[bo:da:bo:]}}\ \ \text{\textipa{[la:no:la:]}}&&\cdots \\
    sharp&: \text{\textipa{[ki:tEki:]}}\ \ \text{\textipa{[zEpa:zE]}}\ \ \text{\textipa{[\textteshlig a:ti:\textteshlig a:]}}&&\cdots
\end{align*}
Finally, we utilized commercial text-to-speech models\footnote{https://cloud.google.com/text-to-speech} to generate audio. 
With each sequence synthesized with four distinct speaker identities, we yield two sets of audio samples, $\mathcal{S}_{\text{round}}$ and $\mathcal{S}_{\text{sharp}}$, each containing 1944 samples.

\section{Evaluation Method}

To examine the presence of sound symbolism in the inherent knowledge of a pre-trained audio-visual model, we adopt a non-parametric approach to probe the model in a zero-shot setting. More specifically, we calculate a geometric score $s_g$ and a phonetic score $s_p$ using the synthesized dataset, as similar to~\cite{alper2023kiki}\footnote{Our evaluation framework is conceptually similar to that of~\cite{alper2023kiki}. However, it's important to highlight that the scores in~\cite{alper2023kiki} are based solely on single-modality (text) inputs, focusing on the semantics implicitly exhibited in the surface forms of pseudowords. In contrast, our approach uses both audio and visual stimuli to probe the model, aiming to explore semantic similarity across multiple modalities.}.
Our intuition is to identify a one-dimensional semantic direction within the model's shared embedding space that best discriminates between sharp and round attributes. We then project query embeddings onto this direction to obtain a score representing their degree of association with each attribute.
In each pre-trained model, there exists two modules, $\mathrm{F}_a\left( \cdot \right)$ and $\mathrm{F}_i\left( \cdot \right)$, which encode audio and images, respectively, into a shared embedding space.

\noindent
\textbf{Geometric Score}: 
We start with extracting the embeddings of ``round'' and ``sharp'' images $\mathcal{I}_{\text{round}}$ and $\mathcal{I}_{\text{sharp}}$ from the pre-trained model and identify the semantic direction in interest $\mathbf{w}_{g}$ in the embedding space:
\begin{equation}
  \label{eq:geometric_p}
  \mathbf{w}_{g}=\frac{1}{|\mathcal{I}_{\text{round}}|}\sum_{i\in \mathcal{I}_{\text{round}}}{\mathrm{F}_i\left( i \right)} - \frac{1}{|\mathcal{I}_{\text{sharp}}|}\sum_{i\in \mathcal{I}_{\text{sharp}}}{\mathrm{F}_i\left( i \right)}
\end{equation}
Then, for each sound $s \in  \mathcal{S}_{\text{round}}\cup\mathcal{S}_{\text{sharp}}$, we define the geometric score as the cosine similarity between its extracted embedding $\mathrm{F}_a\left( s \right)$ and $\mathbf{w}_{g}$.
\begin{equation}
  \label{eq:geometric_score}
  \text{GeometricScore}(s) = \frac{ \mathrm{F}_a\left( s \right) \cdot  \mathbf{w}_{g}} {\lVert \mathrm{F}_a\left( s \right) \rVert \lVert \mathbf{w}_{g} \rVert}
\end{equation}
In other words, we probe the sound $s$ by projecting its embedding onto a semantic axis to determine its degree of association with "round" and "sharp" visual attributes. While we use the geometric score to analyze the sounds of nonwords, this score can also be applied to real images and real words.

\noindent
\textbf{Phonetic Score}: The phonetic score is calculated similarly to the geometric score, but using round and sharp sounds instead of images. Specifically, we feed "round" and "sharp" sounds ($\mathcal{S}_{\text{round}}$ and $\mathcal{S}_{\text{sharp}}$) into our model to obtain two groups of embeddings and calculate the semantic direction:
\begin{equation}
  \label{eq:phonetic_p}
  \mathbf{w}_{p}=\frac{1}{|\mathcal{S}_{\text{round}}|}\sum_{s\in \mathcal{S}_{\text{round}}}{\mathrm{F}_a\left( s \right)} - \frac{1}{|\mathcal{S}_{\text{sharp}}|}\sum_{s\in \mathcal{S}_{\text{sharp}}}{\mathrm{F}_a\left( s \right)}
\end{equation}
Then for each image $i$, we obtain the phonetic score by calculating the cosine similarity between its extracted embedding $\mathrm{F}_i\left( i \right)$  and $\mathbf{w}_{p}$ :
\begin{equation}
  \label{eq:phonetic_score}
  \text{PhoneticScore}(i) = \frac{ \mathrm{F}_i\left( i \right) \cdot  \mathbf{w}_{p}} {\lVert \mathrm{F}_i\left( i \right) \rVert \lVert \mathbf{w}_{p} \rVert}
\end{equation}

\noindent After obtaining the phonetic and geometric scores, we investigate using them to indicate the presence or absence of the kiki/bouba effect in a model. If these scores can effectively distinguish either sounds or images from two predefined groups (round and sharp), it indicates that the sound symbolism pattern is embedded in the model's inherent knowledge. Since these scores are unnormalized, we report ROC-AUC (Receiver Operating Characteristic -- Area Under the Curve) and Kendall's rank correlation coefficient.
ROC-AUC measures the ability of binary classification models to distinguish between positive and negative classes, while Kendall's rank correlation coefficient measures the ordinal association between two variables.
One significant benefit of these metrics is that they are threshold-agnostic, allowing for a thorough assessment of the model's discriminative ability.

\section{Experiments}
\subsection{Pre-trained Audio-Visual Models}
In our experiments, we include eight pre-trained audio-visual models to evaluate the presence of sound symbolism. We briefly introduce the methodologies and pre-training datasets of these models in the following sections and provide an overview in Table~\ref{tab:models}. It is important to note that there are significant domain mismatches between the pre-training datasets of these models, which range from spoken image captions and general audio events to human action recordings and egocentric videos. The data domain might be a crucial factor in establishing sound symbolism patterns. Intuitively, we expect that models trained on spoken image captions are the most likely to exhibit sound symbolism due to the synergy of their pre-training with the human language acquisition process. However, we also aim to determine whether the sound symbolism phenomenon can emerge under weak-linguistic contexts.
Additionally, following~\cite{baltruvsaitis2018multimodal}, the learning objectives of these models can be roughly categorized into learning joint representations, coordinated representations, or both. Joint representations combine the unimodal signals into the same representation space, while coordinated representations process unimodal signals separately but enforce certain similarity constraints on them. We use this categorization to examine whether the pre-training algorithm significantly affects the capture of sound symbolism.We use these models as-is to extract embeddings unless otherwise stated.\footnote{For models that learn joint representations, we extract embeddings by feeding one modality at a time while setting the other modality to zero. For models with coordinated representations, we use their pre-defined single modality encoders.}

\noindent
\textbf{SpeechCLIP}~\cite{shihSpeechCLIP} extends CLIP~\cite{radford2021clip} by adding an extra speech encoder, aligning speech, image, and text within the same embedding space. During pre-training, the model learns to align spoken captions with images through a contrastive learning task. The pre-training dataset of SpeechCLIP is SpokenCOCO~\cite{harwath2016unsupervised}, which contains spoken image captions.


\noindent
\textbf{FaST-VGS}~\cite{peng2022fastvgs} is designed for fast and accurate speech-image retrieval with a single model. 
During its pre-training, the model learns to align speech and image embeddings through a contrastive learning objective. 
The pre-training datasets includes Places Audio~\cite{harwath2015deep}, Flickr8K Audio Caption Corpus~\cite{hsu2021text}, and SpokenCOCO, all of which contain spoken images captions.



\noindent
\textbf{AV-HuBERT}~\cite{shi2022avhubert} is specifically designed for tasks involving audio-visual speech recognition and lip reading. 
During pre-training, parts of the input in both audio and visual modalities are masked, and the model is trained to predict the pre-discovered multimodal cluster assignments for the masked portions.
This joint training helps the model learn the correlation between lip movements and speech sounds. 
The pre-training dataset includes LRS3~\cite{afouras2018lrs3} and VoxCeleb2~\cite{zhu2021deep}, both of which contain videos of people talking with a specialized focus on their profile.

\noindent
\textbf{CAV-MAE}~\cite{gong2023cavmae} first extends single-modality MAE into a audio-visual multi-modality learner, followed by integrating contrastive learning and masked autoencoding into a single framework for enhanced representation. 
This approach simultaneously learns audio-visual pair information and audio-visual correspondence, facilitating the learning of joint and coordinated representations.
AudioSet is used in pre-training.
We select this model to test sound symbolism in audio-visual model that isn't focusing on speech in pretraining.

\noindent
\textbf{MaVIL}~\cite{huang_mavil} learns audio-visual representations by integrating three types of self-supervision: (1) masked raw audio-video reconstruction, (2) inter-modal and intra-modal contrastive learning with masking, and (3) masked contextualized audio-video representation reconstruction via student-teacher learning. The pretraining dataset is AudioSet.

\noindent
\textbf{RepLAI}~\cite{mittal2022learning} is a representation framework designed for egocentric video data by learning from audible interactions within those videos. During pretraining, the model minimizes two types of loss: one that models audio-visual correspondence and another that captures visual state changes caused by audible interactions. The pretraining dataset is Ego4D~\cite{grauman2022ego4d}, which consists of egocentric videos depicting daily human activities and mainly containing object sounds.

\noindent
\textbf{ImageBind}~\cite{Girdhar2023Imagebind} adopts an approach similar to CLIP, learning a joint embedding across six different modalities: images, text, audio, depth, thermal, and IMU data. It achieves this by binding them solely with image-paired data via contrastive loss. This method allows ImageBind to create a unified embedding space where inputs from diverse modalities are effectively aligned and correlated. The pre-training dataset is a combination of five datasets~\cite{radford2021clip,AudioSet,grauman2022ego4d,SUNRGBD,jia2021llvip}.
We include this model to test whether including multiple modalities together help model to improve sound symbolism.

    
   
  

\begin{table}[!t]
\centering
\begin{tabular}{lcc@{\extracolsep{6pt}}p{0.2cm}@{\extracolsep{0pt}}p{0.2cm}p{0.6cm}}
\hline
\multirow{2}{*}{\textbf{Models}} & \multicolumn{2}{c}{\textbf{Repr. Type}} & \multicolumn{3}{c}{\textbf{Data Domain}}   \\ \cline{2-3} \cline{4-6}
                                 & \textbf{Joint}     & \textbf{Coord.}    & \multicolumn{1}{c}{\textbf{S}}            & \multicolumn{1}{c}{\textbf{A}}            & \multicolumn{1}{c}{\textbf{AV-S}}           \\ 
\hline
SpeechCLIP*~\cite{shihSpeechCLIP}                       &                    & $\checkmark$       & \multicolumn{1}{c}{$\checkmark$} &              &              \\
FaST-VGS~\cite{peng2022fastvgs}                         &                    & $\checkmark$       & \multicolumn{1}{c}{$\checkmark$} &              &              \\
AV-HuBERT~\cite{shi2022avhubert}                        & $\checkmark$       &                    &              &              & \multicolumn{1}{c}{$\checkmark$} \\
CAV-MAE~\cite{gong2023cavmae}                          & $\checkmark$       & $\checkmark$       &              & \multicolumn{1}{c}{$\checkmark$} &              \\
MAViL~\cite{huang_mavil}                            & $\checkmark$       & $\checkmark$       &              & \multicolumn{1}{c}{$\checkmark$} &              \\
RepLAI~\cite{mittal2022learning}                           &                    & $\checkmark$       &              & \multicolumn{1}{c}{$\checkmark$} &              \\
ImageBind**~\cite{Girdhar2023Imagebind}                        &                    & $\checkmark$       &              & \multicolumn{1}{c}{$\checkmark$} &              \\
\hline
\end{tabular}
\caption{Overview of pretrained audio-visual models used in our experiments. Based on the taxonomy in~\cite{baltruvsaitis2018multimodal}, we categorize their learning objectives into joint or coordinated representation (or both). We also list the pretraining dataset domains: S for spoken captions, A for general audio events, and AV-S for audio-visual speech. *SpeechCLIP is initialized from CLIP, trained on text-image pairs. **ImageBind's pretraining dataset includes image-paired data across various modalities like text, audio, depth, thermal, and IMU data.}
\label{tab:models}
\end{table}

\subsection{Quantitative Results}
\begin{table}[!t]
  \centering
    \begin{tabular}{l@{\extracolsep{4pt}}rrrr}
    \hline
    \multirow{2}{*}{\textbf{Models}} & \multicolumn{2}{c}{\textbf{Geometric}}                     & \multicolumn{2}{c}{\textbf{Phonetic}}                      \\
    \cline{2-3}
    \cline{4-5}
                           & \multicolumn{1}{c}{\textbf{AUC}} & \multicolumn{1}{c}{$\tau$} & \multicolumn{1}{c}{\textbf{AUC}} & \multicolumn{1}{c}{$\tau$} \\
    \hline
    \multicolumn{3}{l}{\textit{vision-language models}} \\
    CLIP~\cite{radford2021clip}	& 0.70                    & 0.28                    & 0.83                    & 0.48                    \\
    ImageBind\textsubscript{text}~\cite{Girdhar2023Imagebind}	& 0.59                    & 0.13                    & 0.70                     & 0.29                    \\
    BLIP~\cite{li2022blip}   & \textbf{0.84}                    & \textbf{0.49}                    & \textbf{0.87}                    & \textbf{0.54}                    \\
    \hline
    \hline
    \multicolumn{3}{l}{\textit{audio-visual models}} \\
    SpeechCLIP~\cite{shihSpeechCLIP}	& \textbf{0.73}                    & \textbf{0.32}                    & \textbf{0.87}                   & \textbf{0.54}                    \\
    FaST-VGS~\cite{peng2022fastvgs}		& \textbf{0.71}                    & \textbf{0.30}                    & \textbf{0.84}                    & \textbf{0.49}                    \\
    AV-HuBERT~\cite{shi2022avhubert}	& 0.47                    & -0.05                   & 0.42                    & -0.12                   \\
    CAV-MAE~\cite{gong2023cavmae}	    & \underline{0.30}                    & \underline{-0.29}                   & \underline{0.31}                    & \underline{-0.28}                   \\
    MAViL~\cite{huang_mavil} & \underline{0.35}                    & \underline{-0.21}                   & 0.48                    & -0.03                   \\
    RepLAI~\cite{mittal2022learning} & 0.57                    & 0.10                    & 0.59                    & 0.13                    \\
    ImageBind\textsubscript{audio}~\cite{Girdhar2023Imagebind}	& 0.68                    & 0.25                    & 0.74                    & 0.35                    \\
   \hline
   \hline
   (Random) & 0.50 & 0.00 & 0.50 & 0.00 \\
   \hline
  \end{tabular}
  
  \caption{Classification performance for both geometric score and phonetic score. AUC represents ROC-AUC (Receiver Operating Characteristic -- Area Under the Curve) and $\tau$ indicates Kendall's rank correlation coefficient.
  }
  \label{tab:scores}
\end{table}
Table~\ref{tab:scores} presents the classification performance of each model using geometric and phonetic scores. For comparison, we also include results from several text-based vision-language models, where the audio input is replaced with text forms (pseudowords) using a predefined phoneme-to-grapheme mapping. The embeddings are extracted using the text encoder of these models instead.

Overall, models that learn coordinated representations achieve better classification performance than those with joint representations (or both). Specifically, SpeechCLIP, FaST-VGS, and ImageBind can significantly distinguish sounds and image classes better than chance. This indicates that sounds in $S_{\circ}$ are more likely to be closer to the visual stimuli of "round" in the embedding space, and vice versa, providing strong evidence for the presence of a sound symbolism pattern in these models. Interestingly, models that learn both joint and coordinated representations show a slightly reversed trend compared to the human sound symbolism pattern, though the reason for this remains unclear.

Furthermore, models pre-trained on spoken image captions generally achieve better classification performance than those pre-trained on general audio events. This finding aligns with evidence of sound symbolic trends in the basic lexicon of the language~\cite{article, monaghan2014arbitrary}. Yet, the result from ImageBind suggests that the sound symbolism pattern might still be facilitated by learning in a weak-linguistic context. Additionally, we observe that audio-visual models tend to exhibit a stronger sound symbolism pattern compared to their vision-language model counterparts (e.g., SpeechCLIP versus CLIP). One possible explanation is that the sound-meaning association is inherently linked to the phonetic and articulatory characteristics of sounds~\cite{Synaesthesia, ShapeOfBoubas}, which are less profound in text forms, making such associations harder to learn.

Lastly, AV-HuBERT, which is trained on audio-visual speech, shows few sound symbolism patterns, exhibiting near-random classification performance. This highlights the importance of interaction with real-world visual concepts to build cross-modal correspondences, as supported by prior psychological literature~\cite{FletcherVisual}. Additionally, comparing ImageBind and CLIP reveals that training on larger and more diverse datasets is not beneficial for establishing sound-symbolism patterns. Evaluating the more recent vision-language model, BLIP~\cite{li2022blip}, demonstrates an enhanced sound-symbolism effect. This finding suggests a potential link between sound symbolism and language understanding capability, as better language understanding performance implies better discrimination between the semantics of opposite concepts across modalities and better alignment for the semantics of similar concepts. We leave the exploration of this potential linkage for future work.

\begin{figure}
    \centering
    \includegraphics[width=0.9\linewidth,trim={0cm 2.2cm 0 0.1cm},clip]{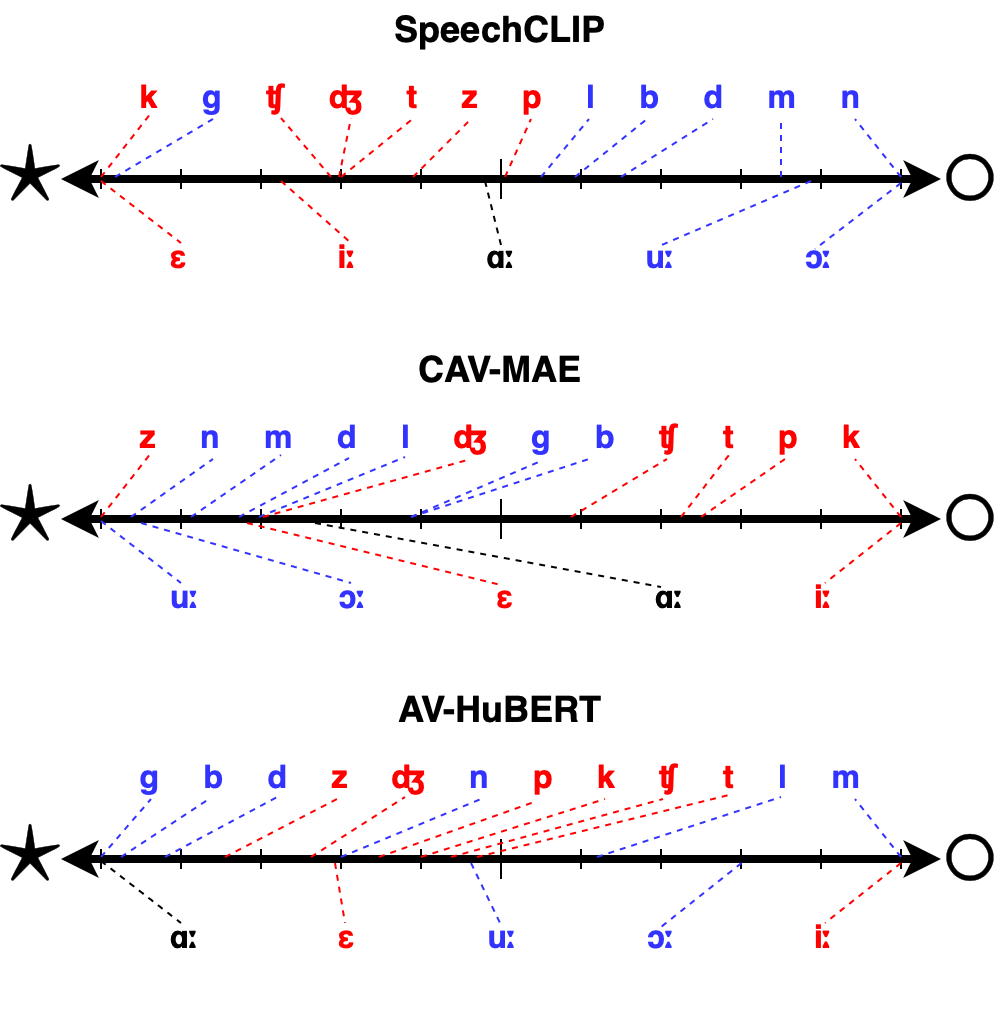}
    \caption{Phones sorted by average geometric score grouped by the first syllable of the sounds.
    The colors indicates ground-truth association of each phone. (\textcolor{blue}{blue} and $\circ$ refer to round group, while \textcolor{red}{red} and $\star$ represent sharp).
    Consonants and vowels are displayed on separate scales but are positioned absolutely to each other within each scale.
    }
    \label{fig:visualization}
    \vspace{-10pt}
\end{figure}

\subsection{Qualitative Results}

Beyond quantitative results, we also provide a qualitative analysis for a more interpretable view of our findings. Following a similar approach to~\cite{alper2023kiki}, we group the sounds by their first grapheme and compute the average geometric score for each consonant and vowel. The results are visualized in Fig~\ref{fig:visualization}. Due to space limitations, we only show results for several models here.

We observe that SpeechCLIP effectively distinguishes phones of different groups and aligns well with human preferences~\cite{mccormick2015sound}. In contrast, CAV-MAE and AV-HuBERT show slightly reversed tendencies and near-random patterns, respectively. These qualitative insights complement our quantitative results, providing a more comprehensive understanding of how sound symbolism manifests in audio-visual models.

\section{Conclusion}
In this work, we investigate the presence of the kiki-bouba effect in pre-trained audio-visual models. Using a non-parametric approach in a zero-shot setting with specialized data, we probe the inherent knowledge of these models. Our results reveal a significant correlation between the models’ outputs and established patterns of sound symbolism in some audio-visual models. Specifically, audio-visual models trained on spoken image captions exhibit kiki-bouba effect patterns similar to those observed in humans. In contrast, models trained with general audio events show random or reversed tendencies. Additionally, using speech input demonstrates a more pronounced sound symbolism effect than using text forms. Our findings align with prior psychological literature, contribute valuable insights regarding the non-arbitrariness of language in linguistic theory.

\section{Limitations}
This study specifically investigates the kiki-bouba effect, a well-known case of sound symbolism. 
While this focus allows for a detailed exploration of this phenomenon, we acknowledge the existence of other forms of sound symbolism across human languages. 
Future research will aim to expand our investigations to include various forms such as phonesthemes~\cite{firth1968tongues}, ideophones~\cite{dingemanse2019ideophone}, and magnitude symbolism~\cite{winter2021size}, enabling a more comprehensive understanding of sound symbolism phenomenon in audio-visual models.

Additionally, our current analysis is limited to audio-visual models pre-trained on English-language corpora. In subsequent studies, we intend to evaluate models pre-trained on corpora from diverse languages, which could provide additional insights into the language-agnostic properties of sound symbolism, as suggested by previous studies~\cite{kohler1967gestalt, bremner2013bouba}.

Lastly, there is a potential to further explore practical applications of sound symbolism by linking it to other audio-visual tasks. 
Such applications include designing a sound-symbolism related training objective or studying the relation between audio-visual understanding tasks~\cite{tseng2023avsuperb} with a model's sound symbolism ability.
Establishing these connections could enhance the utility of sound symbolism in broader computational and cognitive science contexts.

\bibliographystyle{IEEEbib}
\bibliography{strings,refs}

\end{document}